\documentclass[10pt,letterpaper]{article}

\usepackage{cogsci}
\usepackage{pslatex}
\usepackage{apacite}
\usepackage{graphicx}
\usepackage{subfigure}
\usepackage{microtype}
\usepackage{natbib}
\usepackage{booktabs}
\usepackage{hyperref}

\usepackage{url}
\usepackage{verbatim}
\usepackage{amsfonts}
\usepackage{amsmath}
\usepackage{float}

\title{Emergence of Structured Behaviors from Curiosity-Based Intrinsic Motivation}
 
\author{{\large \bf Nick Haber, Damian Mrowca, Li Fei-Fei, Daniel L. K. Yamins} \\
{\large \bf ( [nhaber, mrowca, feifeili, yamins]@stanford.edu )} \\
  Department of Psychology and Computer Science, Stanford University, Stanford, CA 94305, USA}

\begin{document}

\maketitle

\begin{abstract}
Infants are experts at playing, with an amazing ability to generate novel structured behaviors in  unstructured environments that lack clear extrinsic reward signals.  We seek to replicate some of these abilities with a neural network that implements curiosity-driven intrinsic motivation.  Using a simple but ecologically naturalistic simulated environment in which the agent can move and interact with objects it sees, the agent learns a world model predicting the dynamic consequences of its actions.  Simultaneously, the agent learns to take actions that adversarially challenge the developing world model, pushing the agent to explore novel and informative interactions with its environment.  We demonstrate that this policy leads to the self-supervised emergence of a spectrum of complex behaviors, including ego motion prediction, object attention, and object gathering.  Moreover, the world model that the agent learns supports improved performance on object dynamics prediction and localization tasks.  Our results are a proof-of-principle that computational models of intrinsic motivation might account for key features of developmental visuomotor learning in infants. 
\textbf{Keywords:} 
Development learning, Curiosity, Neural Network Models
\end{abstract}

\section{Introduction}
\label{introduction}
Within the first year of life, humans exhibit a wide range of interesting, apparently spontaneous, visuomotor behaviors --- including navigating their environment, seeking out and attending to objects, and engaging physically with these objects in novel and surprising ways \citep{fantz_visualexperienceininfants, twomey_curiositybasedlearning, hurley2010_influenceofpets, hurley2015_petexposure, goupil_infantsask, begus_infantslearnwhattheywant, gopnik_scientistincrib}.
In short, young children are excellent at playing, and their ability to make sense of and (re)structure their environments sets them apart from even the most advanced autonomous robots.
Play capacity in this period likely interacts with infants' powerful abilities to understand and model their environment.
By six months of age or younger, infants can orient themselves in a complex environment, account for the presence, number and visual properties of objects they interact with, and have a sense of how these objects behave dynamically \cite{spelke_objectpermanence, stahl_observingunexpected, baillargeon_eightlessons}.

But how exactly do such young children know how to play? And how do such behaviors relate to their world-model building abilities?  
One natural idea is that infants' world-modeling capacities are the result of built-in core systems, including those for e.g. object attention and permanence, self-localization, number sense, and intuitive physics \citep{spelke2007core}.
Once operational, such systems would naturally give the infant a basis on which to make judgments about which sequences of actions would be interesting to perform. 
	
\begin{figure}[ht]
\begin{center}
\includegraphics[width=1.0\linewidth]{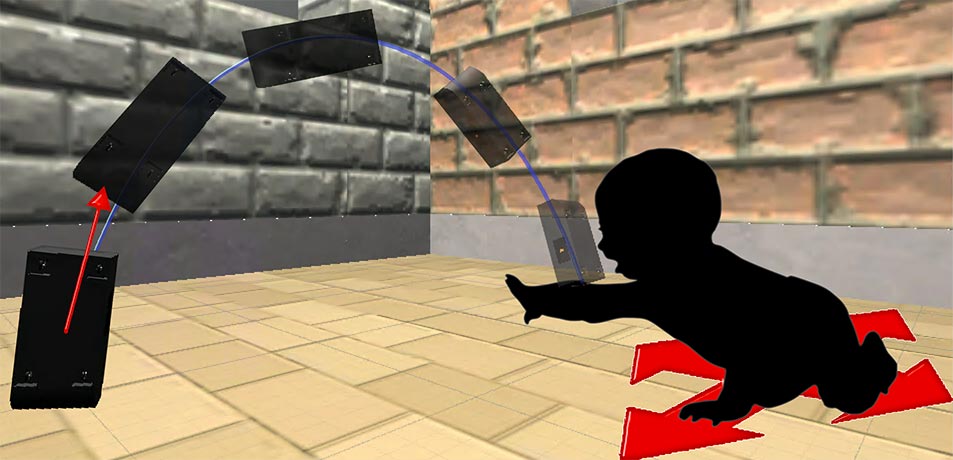}
\end{center}
\vspace{-0.3cm}
\caption{Problem setting. How do agents learn from object interactions while moving around in the physical world?}
\label{fig:setup}
\end{figure}

A related but alternative idea is that the intrinsic motivation of curiosity can itself drive the development of world-model making \citep{schmidhuber_formaltheoryoffun}.
This idea relies on a virtuous cycle in which, by seeking out novel but replicable interactions, the child pushes the boundaries of what its world-model-prediction systems can achieve, giving itself useful data on which to improve and develop these systems. 
As world-modeling capacity improves, what used to be novel becomes old hat, and the cycle starts again.
Related to the conception of the ``scientist in the crib'' \citep{gopnik_scientistincrib}, in this account, aside from being fun, play behaviors may be extremely useful and highly structured, driving the self-supervised learning of a variety of representations underlying sensory judgments and motor planning capacities \citep{selfsupervision_pose}.
    
Building on recent work in artificial intelligence \citep{schmidhuber_formaltheoryoffun, berkeley_mario, kulkarni_hierarchical, jaderberg2016reinforcement} we make a computational model of an agent driven by curiosity-based intrinsic motivation. 
We present a simple simulated interactive environment in which an agent can move around and physically act on objects it sees (Fig. \ref{fig:setup}). 
In this world, interesting interactions are sparse unless actively sought after. 
We then describe a neural network architecture through which the agent learns a world model that seeks to predict the consequences of its own actions.  
In addition, as the agent optimizes the accuracy of its world model, a separate neural network simultaneously learns an agent action policy that seeks to take actions that adversarially challenge the current state of its world model.   
We demonstrate that this architecture stably engages in the virtuous reinforcement learning cycle described above, spontaneously learning to understand self-generated ego motion and to selectively pay attention to, localize, and interact with objects, without having to have any of these concepts built in. 
This learning occurs through emergent self-curricularization in which new capacities arise at distinct ``developmental milestones.''
This work is computational proof-of-principle of how intrinsic motivation could drive aspects of visuomotor learning and play in infants and young children, and of how more flexible intrinsically motivated autonomous robots might be constructed. 

\begin{figure*}[t!]
\begin{center}
\includegraphics[width=\textwidth]{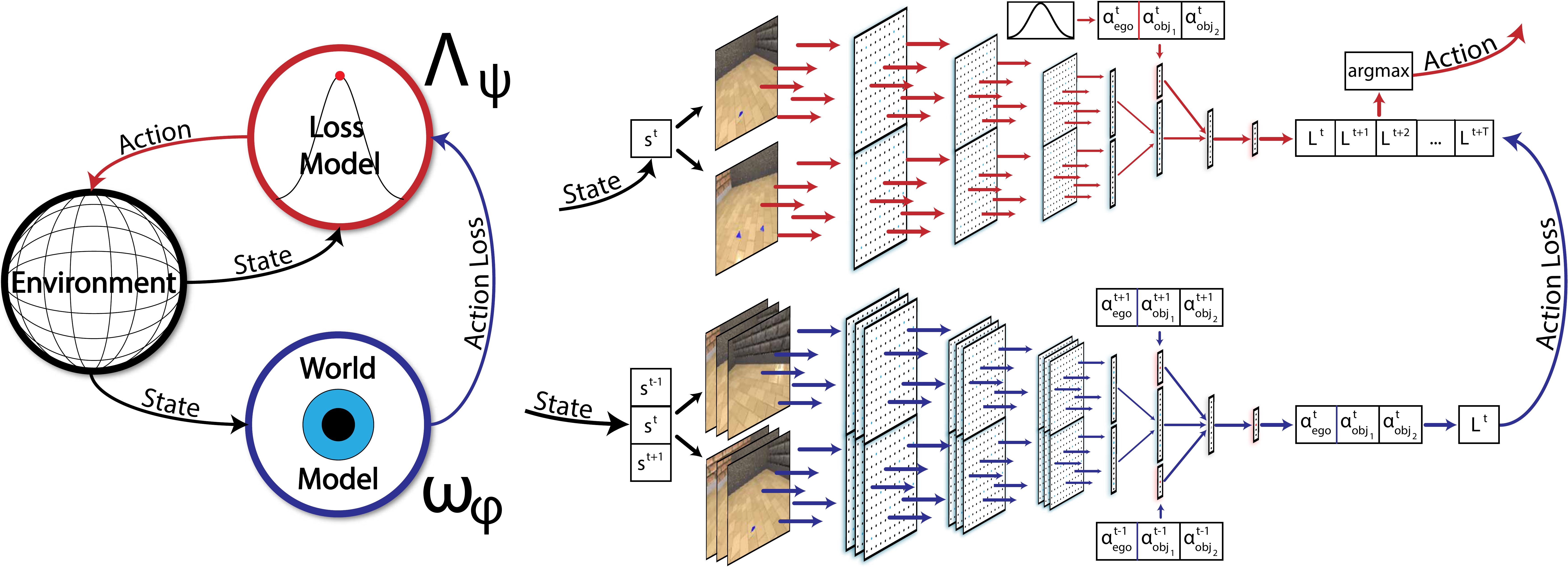}
\end{center}
\vspace{-0.3cm}
\caption{Self-supervised curiosity model. We train a dynamical world model (blue), while simultaneously learning a loss model (red) that predicts the world model's loss to choose actions that lead to novel and surprising events in the environment (black).}
\label{fig:model}
\vskip -0.1in
\end{figure*}

\section{Agent Architecture and Environment}

\label{sec:model}
We place an agent in a physically realistic simulated {\em environment} built in the Unity 3D simulation framework. The agent consists of a world model and a loss model. The {\em world model} is tasked to learn dynamics from visuomotor inputs. The {\em loss model} tries to estimate the world model's losses several time steps into the future to choose actions that antagonize the world model's learning. Our self-supervised curiosity system is depicted in Figure \ref{fig:model}. We emphasize that we do not initialize our model with pretrained weights so as to explore what world representation and behaviors emerge from this simple antagonistic setup in a physically embodied environment.

\subsection{Interaction environment}
Our environment consists of a simple square room in which an agent and several objects are initially placed randomly. 
The agent is modeled as an invisible sphere that can move around and, in discrete time steps, receives an RGB image from its forward-facing camera.
In order to model interaction with objects requiring some attention and proximity, the agent can apply forces and torques in all three dimensions to objects that are both in view and within a fixed distance $\delta$.
We refer to a state in which the agent can act on the object as a {\em play state}, and the object as {\em in play}. Although the remaining environment is static, the agent and objects can collide with every  part.

We define a state in the state space $s_t \in S$ to consist of the images captured at times $t$ and $t-1$ by the agent. In state $s_t$, the agent specifies an action $a_t \in A$, which leads to the next state $s_{t+1}$. The action space $A \in \mathbb{R}^{2+6N}$ is continuous. The first $2$  dimensions specify ego motion, restricting agent movement to forward/backward motion $v_{fwd}$ and horizontal planar rotation $v_{\theta}$.
The remaining $6N$ dimensions specify the forces $f_x, f_y, f_z$ and torques $\tau_x, \tau_y \tau_z$ applied to $N$ objects sorted from the lower-leftmost to the upper-rightmost object relative to the agent's field of view. This representation is unambiguous as objects can only be acted on when in view, and if $k < N$ objects are in play, the environment accepts all $2 + 6N$-tuples but only applies the leftmost $k$ force-torque pairs. All coordinates are bounded by constants and normalized to $1$.

\subsection{World model} 
\label{sec:world_model}
Given a slice of history $h_t = (s_{t - k_b}, a_{t - k_b} \ldots s_{t + k_f}, a_{t + k_f}) \in H$, we can describe a {\em generalized dynamical problem} by an input map $\xi : H \rightarrow X$ and a true-value map $\eta: H \rightarrow Y$ and require the world model (blue in Figure \ref{fig:model}) map $\xi(h)$ to $\eta(h)$ --- regardless of whether this is well-defined. Let $\omega$ denote this world model, so that $\omega(\xi(h)) \in Y$. For each prediction, a loss $L_{wm}(\omega(\xi(h)), \eta(h))$ is incurred. In theory, future prediction makes an attractive dynamical problem, with 
$H = \{h = (s_{t - k}, a_{t - k} \ldots s_{t}, a_{t}, s_{t + 1})\}$, $\xi(h) = (s_{t - k}, a_{t - k} \ldots s_{t}, a_{t}),$ and $\eta(h) = s_{t + 1}$. In practice, we find {\em inverse dynamical prediction} useful --- filling in a missing action, instead of predicting the future --- and concretely in our case, $H = \{(a_{t - 2}, s_{t-1}, a_{t - 1}, s_{t}, a_t, s_{t+1}, a_{t+1}, s_{t+2})\}$ with $\xi(h)$ excluding $a_t$ and $\eta(h) = \zeta(a_t)$. Here $\zeta$ zeros out the force and torque components not corresponding to objects in play, as they have no observable effect, and bins the action into classes, with $L_{wm}$ the softmax cross-entropy loss. We bin each dimension by $x < -0.1, -0.1 \leq x \leq 0.1,$ and $x > 0.1$. 

We train a convolutional neural network $\omega_\phi$ for this task from scratch with stochastic gradient descent, with randomly initialized parameters $\phi$. We use twelve convolutional layers, with two-stride max pools every other layer and one hidden layer to encode all states into a lower-dimensional latent space with shared weights. The latent states $\{\lambda_{t+i}\}$ concatenated with the given actions $\{a_{t+j}\}$ are then input to a two-layer MLP to predict $a_t$.

\subsection{Loss model}

The agent's goal is to antagonize the world model, so if it could predict the loss incurred at future time steps as a function of its options, a policy could be made. In practice, we do this explicitly, except predicting only a discretization of the loss for ease of training. Given $s_t$ and a proposed next action $a$, the loss model $\Lambda$ (red in Figure~\ref{fig:model}) predicts a probability distribution over $C_l$ discrete (via thresholding) classes of world model loss for a set number $T$ of future time steps. It is penalized with a softmax cross-entropy loss 
$L_{lm}(\Lambda(s_t), (L_{wm}(\eta_(h_{t + s}), \omega_\phi(\xi(h_{t}))) \ | \ s \in 1 \ldots T)) $. We use a separate convolutional neural network $\Lambda_\psi$ with parameters $\psi$, with twelve convolutional layers with two-stride max pools every other layer and one hidden layer to encode the state which is then concatenated with the proposed $a_t$. This representation is then used as input to a two-hidden-layer MLP to infer the prediction. Note that all future losses aside from the first one, depend not only on the state of the world model, but also on future actions taken, and the loss model hence needs to predict in expectation over future policy. The loss predictions are usefully interpreted as {\em loss prediction maps} $\Lambda_{s_t}(a)$ on action space given a current state $s_t$ as depicted in Figure \ref{fig:lossmap_examples}.




\subsection{Action policy}
Given the loss prediction model, the agent can use a simple mechanism to choose its actions. The loss model provides, given $s_t$ and a proposed next action $a$, $T$ probability distributions 
${\bf p} (k \ | \ s_t, a) = (p^1_{lm}(k \ | \ s_t) \ldots p^T(k \ | \ s_t, a_t)), k \in \{1 \ldots C_l\}.$
Given a real-valued function $\sigma$ of these $T$ probability distributions, we can define our policy $\pi(a | s_t)$ as a distribution
\begin{equation} \label{eq:policy}
\pi(a| s_t) \sim \exp(\beta \sigma(\bf{p} (k \ | \ s_t, a)),
\end{equation}
with hyperparameter $\beta$. For our purposes, taking expectation in loss class $k$ and sum over all time steps is sufficient.
In practice, we execute our policy by evaluating $\sigma$ for $K$ uniform random samples in $A$. We then sample from a $K$-way discrete distribution with probabilities proportional to equation \eqref{eq:policy}.
In choosing this policy mechanism, we opt to start with a na\"ive approach over using more sophisticated reinforcement learning standards in an effort to focus on studying the structure of the proposed self-supervision signal. By explicitly predicting loss separately for several time steps into the future, our results admit easy visualization and interpretation.

\section{Experiments}
\label{sec:experiments}
We randomly situate the agent and up to two objects in a square 10x10 unit room with play distance $\delta = 2$. The agent trains on 16 blue objects with different shapes, i.e. cones, cylinders, cuboids, pyramids, and spheres of varied aspect ratios. We reinitialize the play scene every $8,000$ to $30,000$ steps.

We first place the agent with one object in the room and show that it learns to predict ego motion, and attend, localize and navigate towards objects by evaluating the world model's training loss, the agent's play state frequency, and the inverse dynamics prediction performance on a fixed validation set. In a second experiment we increase the number of objects to two, and demonstrate that the agent learns to gather both objects and further prefers to play with two objects over one object by looking at the frequency of 1 and 2 object play states and object-agent distances. Overarching these results is the observation that these behaviors emerge in a specific order akin to developmental milestones.

We compare our learned world model with curious policy with $T=40$ (LW/CP-40) against a baseline with a world model with fixed random weights following a random policy (RW/RP) and a baseline where the world model weights are learned with a random policy (LW/RP).

\begin{figure*}[ht]
\begin{center}  
	\includegraphics[width=\linewidth]{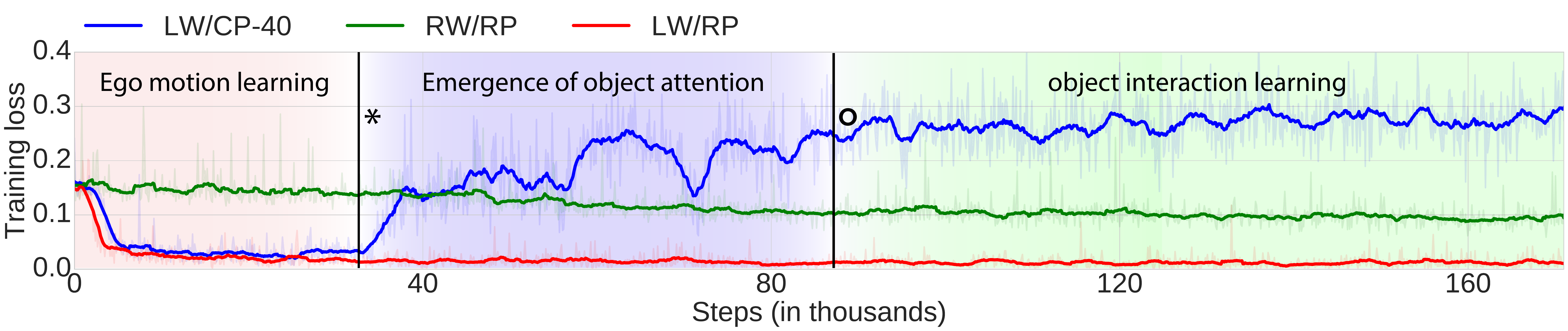} \\
    (a) World model training loss
	\begin{tabular}[c]{@{\hspace{-0.1pt}}ccc}
		\includegraphics[width=0.32\linewidth]{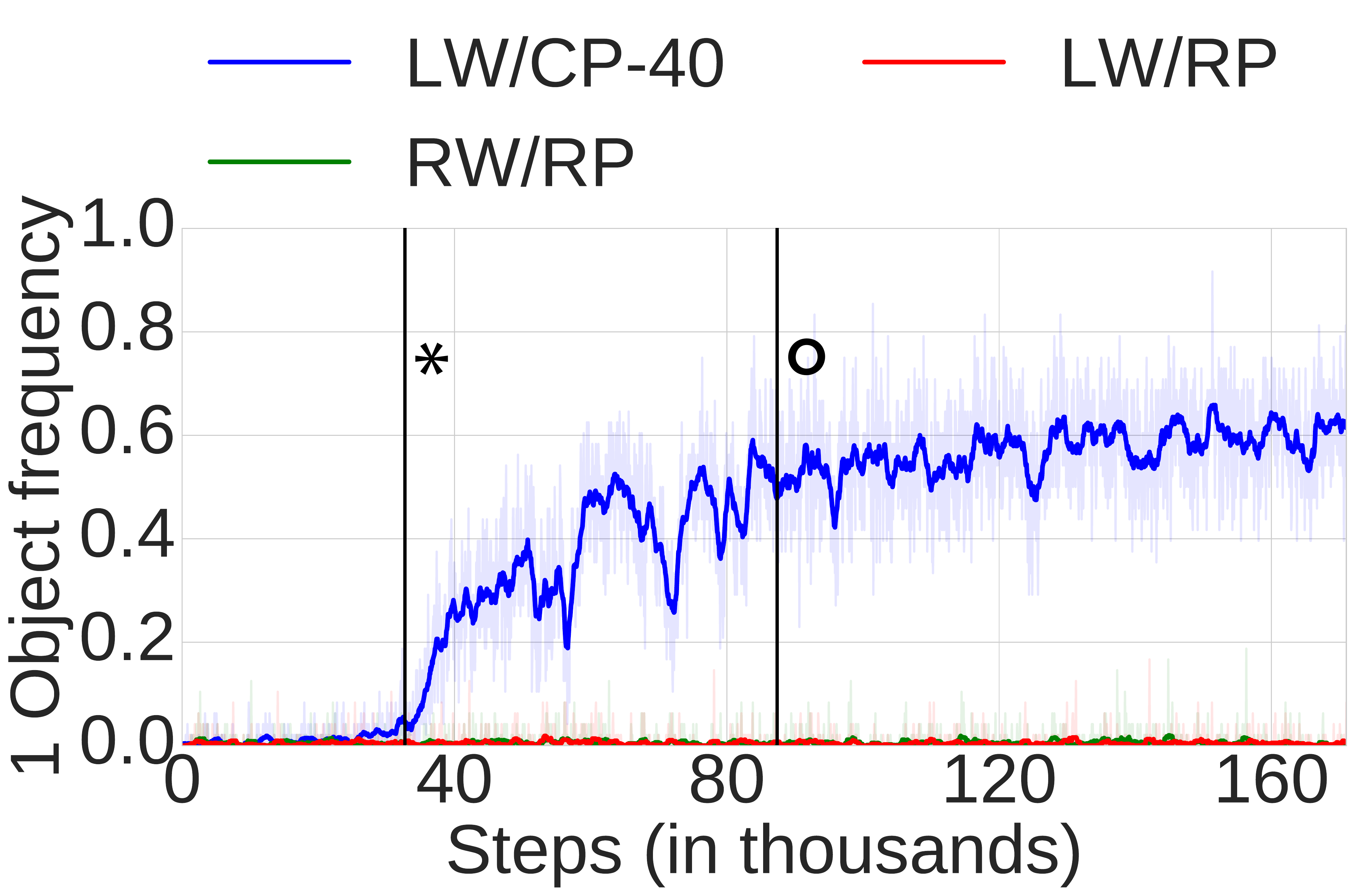}
        &
        \includegraphics[width=0.32\linewidth]{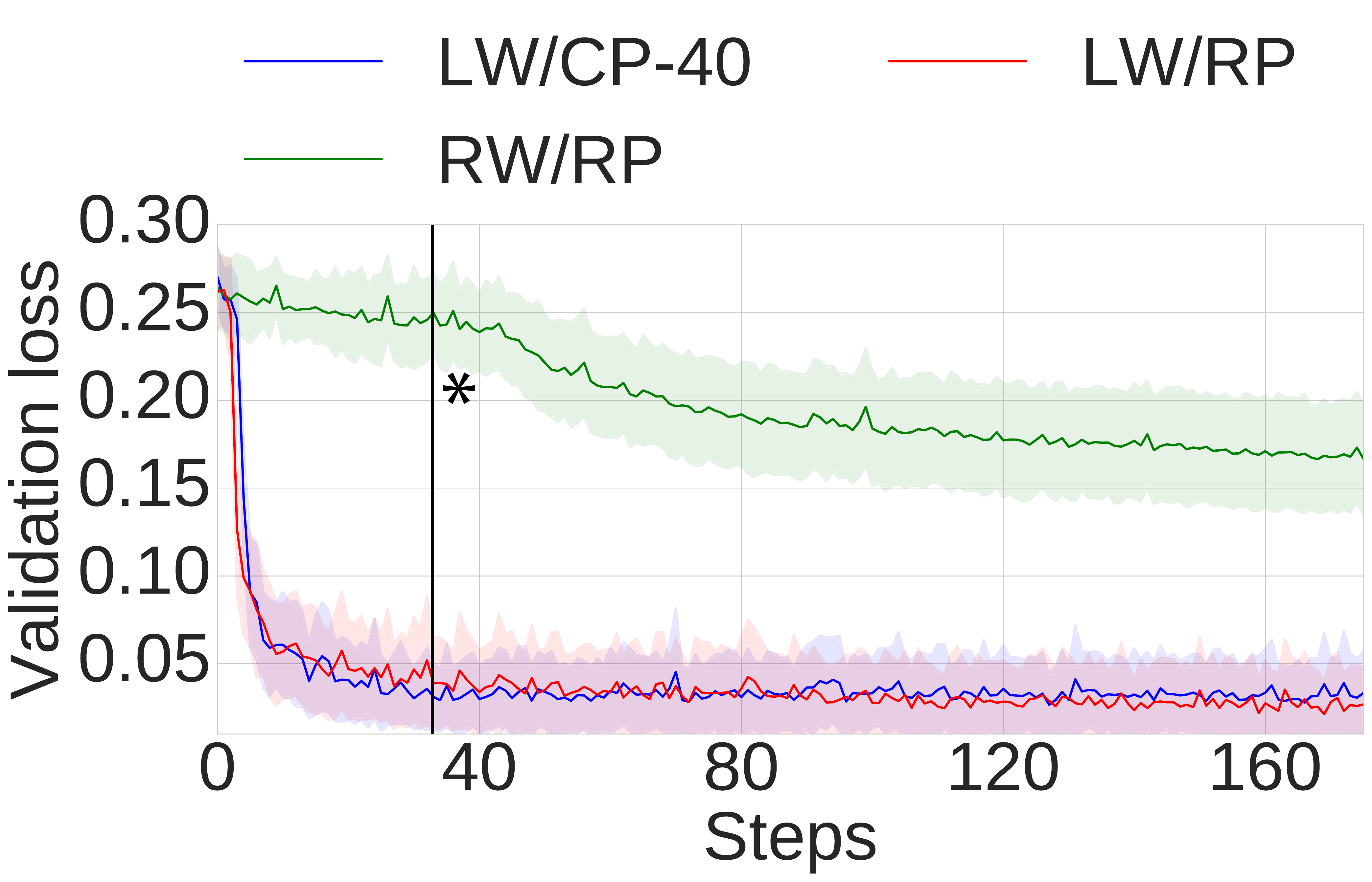} 
        &
        \includegraphics[width=0.32\linewidth]{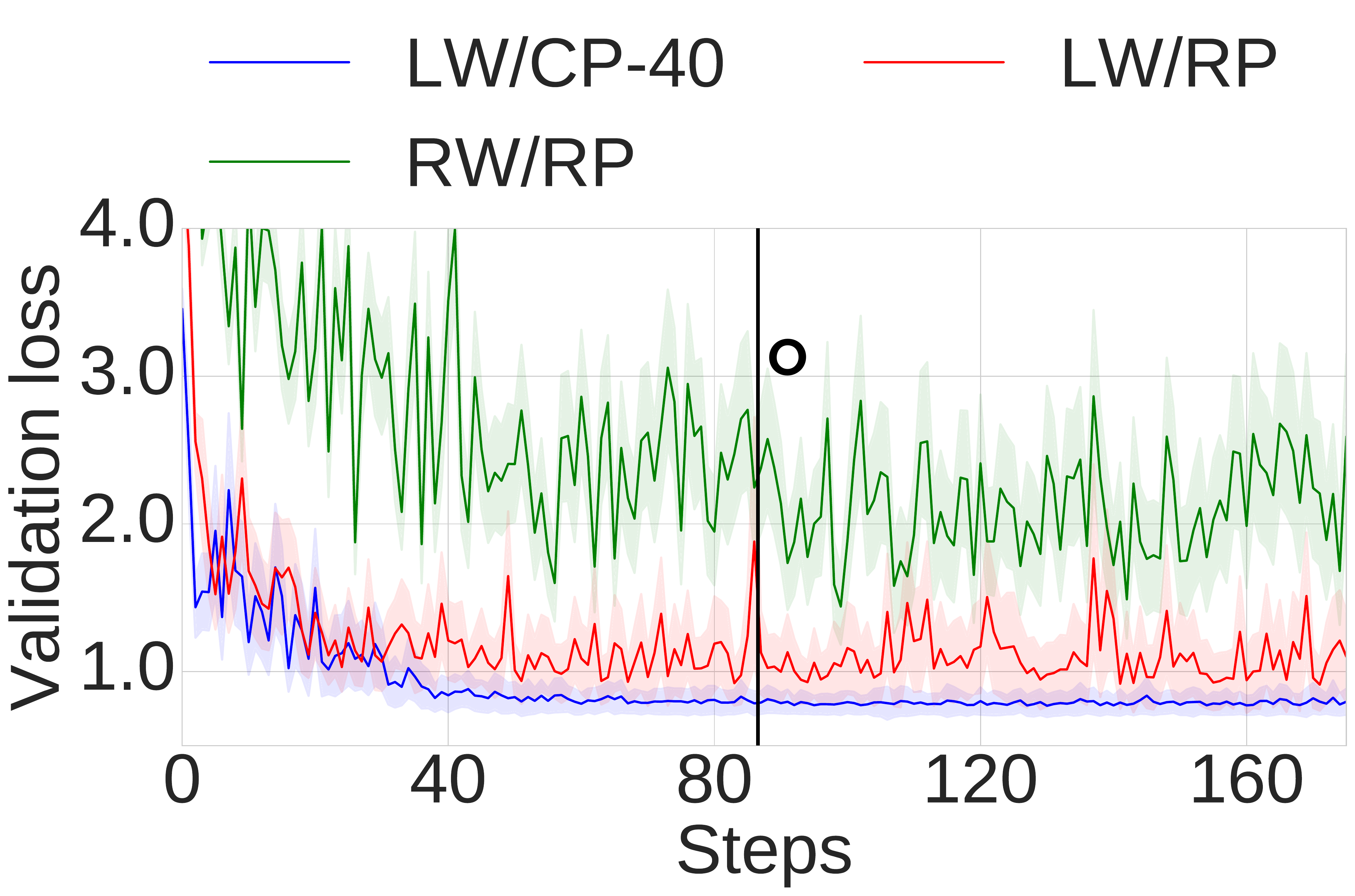}
        \\
        (b) 1 object play frequencies
        &
        (c) Ego motion validation loss
        &
        (d) 1 object interaction validation loss
        \\
	\end{tabular} \\
\end{center}
\vspace{-0.3cm}
\caption{1 object experiments. Learned world model with curious policy (LW/CP-40) is compared against learning a world model following a random policy (LW/RP) and a world model with random weights with a random policy (RW/RP). (a) World model cross-entropy loss during training. (b) Object play state frequency in \%. (c) Ego motion prediction cross-entropy loss on held out validation set. (d) 1 object interaction prediction cross-entropy loss on held out validation set.}
\label{fig:1obj_experiments}
\vskip -0.1in
\end{figure*}

\subsection{Ego motion learning}

Figure \ref{fig:1obj_experiments} (a) shows the training loss curves of LW/CP-40 and the baselines. RW/RP does not learn well since most of its weights are fixed to be random. LW/RP quickly converges to a low value as it learns from a constant random distribution without an antagonistic policy. The LW/CP-40 loss dips before increasing as the loss model first needs to learn which actions lead to higher loss before being able to antagonize the world model effectively.
This first dip in loss corresponds to the world model learning ego motion. The ego motion error reported in Table \ref{table:results} is close to the error reached at this point. 

\subsection{Emergence of object attention}

As the LW/CP-40 loss increases after the initial ego motion dip the agent starts to attend to objects which is reflected in an increase of object interactions as shown in Figure \ref{fig:1obj_experiments} (b). At the final stage, the agent interacts with the object about $60 \%$ of the time which indicates that it learns to localize and attend to the object (Table \ref{table:results}). At the same time, the increasing world model loss shows that these object interactions are much harder to predict than ego motion. The baselines almost never interact with the object and thus experience lower ego motion losses.

\subsection{Improved inverse dynamics prediction}

We evaluate the inverse dynamics prediction performance on a held out validation set gathered from the environment while following a random action policy. To measure the models' performance on predicting ego motion and object actions separately, we divide the validation set into two sets. The first set contains all examples in which the agent is in a play state. The second set consists of all remaining examples. As we can see in Figure \ref{fig:1obj_experiments} (c) and in Table \ref{table:results}, LW/CP-40 and LW/RP perform well on predicting ego motion as no antagonistic policy is necessary to encounter ego motion. However, our policy outperforms the baselines on predicting object interactions by a significant margin showing that focusing on object interactions does indeed improve inverse dynamics prediction performance as seen in Figure \ref{fig:1obj_experiments} (d) and Table \ref{table:results}.

\begin{table}[ht]
\caption{Performance comparison. Learned world model with a curious policy (LW/CP-40) is compared against learned world model with random policy (LW/RP) and random world model with random policy (RW/RP). Ego motion ($v_{fwd}$, $v_\theta$) and interaction ($f,\tau$) accuracy in \% is compared for play and non-play states. Object frequency and presence are measured in \% and localization in mean pixel error.}
\label{table:results}
\vspace{-0.1cm}
\begin{center}
\begin{small}
\begin{tabular}{lccc}
\toprule
\textbf{Task} & \textbf{RW/RP} & \textbf{LW/RP} & \textbf{LW/CP-40} \\
\midrule
Play $v_{fwd}$ accuracy & 61.8 & 84.5 & \textbf{93.6} \\
Play $v_\theta$ accuracy & 78.8 & 95.2 & \textbf{97.9} \\
Play $f,\tau$ accuracy & 20.8 & 39.8& \textbf{44.4}\\
Non-play $v_{fwd}$ accuracy & 59.2 & \textbf{94.2} & 92.3 \\
Non-play $v_\theta$ accuracy & 75.0 & \textbf{97.9} & \textbf{97.9} \\
Object frequency & 0.2 & 0.4 & \textbf{59.0} \\
\midrule
Presence error & 3.9 & 1.2 & \textbf{0.6} \\
Localization error [px] & 15.14 & 5.99  & \textbf{4.80} \\
\bottomrule
\end{tabular}
\end{small}
\end{center}
\vskip -0.1in
\end{table}


\subsection{Improved object detection and localization}

To quantify the world model's object presence and localization performance, we train a linear regression/logistic regression with elastic net regularization on features from various world model layers on an offline dataset generated by gathering data online while following a random action policy. Half of the object presence training data contains an object. For the localization experiment, the second image is guaranteed to contain the object. Both training datasets consists of 16,000 image pairs labeled with the object's presence or pixel-wise 2d position of its centroid respectively. The respective validation and test sets comprise 8,000 image pairs each. 
As can be see in Table \ref{table:results} our model outperforms the baselines on the object presence and localization task, indicating that it learns better visual features.

\begin{figure*}[ht!] 
\begin{center}  
 	\begin{tabular}[c]{ccccc}
         \includegraphics[width=0.17\textwidth]{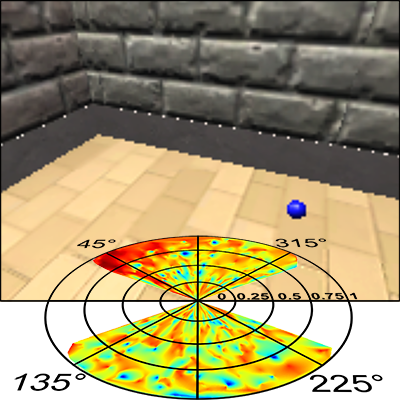}
         &
         \includegraphics[width=0.17\textwidth]{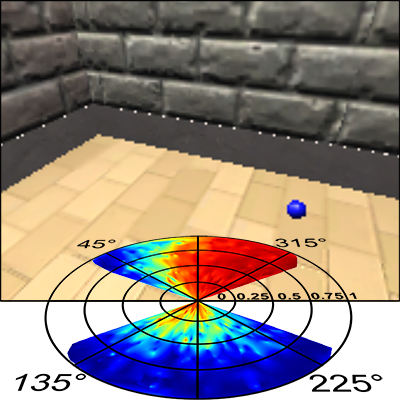}
         &
         \includegraphics[width=0.17\textwidth]{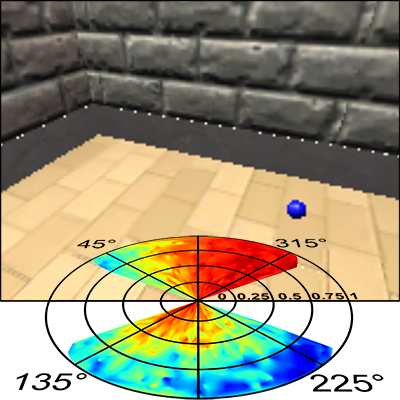}
         &
         \includegraphics[width=0.17\textwidth]{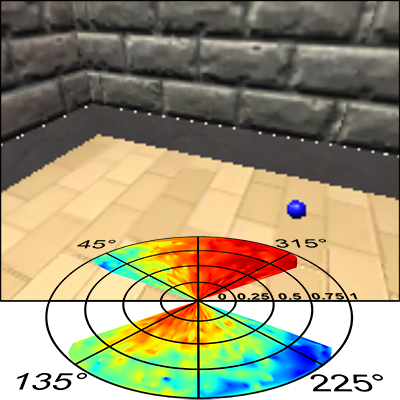}
         &
         \includegraphics[width=0.17\textwidth]{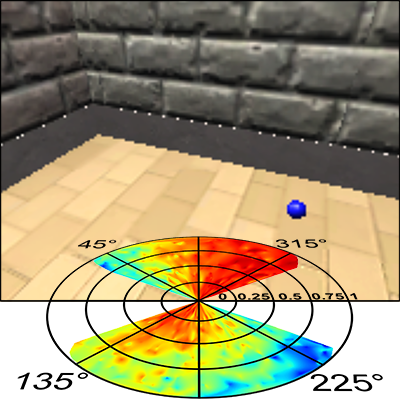}
         \\
         t & t+1 & t+2 & t+3 & t+4 \\
 	\end{tabular}
\end{center}
\vspace{-0.3cm}
\caption{Navigation and planning behavior. The loss model predicts higher loss if the agent turns towards the object. Red colors correspond to high and blue colors to low loss predictions. The center of the heat map corresponds to the agents position.}
\label{fig:lossmap_examples}
\vskip -0.1in
\end{figure*}

\subsection{Navigation and planning}
In addition to object localization, the agent also exhibits navigation and planning abilities. In Figure \ref{fig:lossmap_examples} we give visualizations of loss maps projected onto the agent's position at the respective time. The {\em loss prediction maps} are generated by uniformly sampling $1000$ actions $a$ from the action space $A$, evaluating $\Lambda_{s_t}(a)$ and applying a postprocessing smoothing algorithm. We truncate the figure at five out of the 40 time steps our loss model predicts. The loss maps show the agent predicting a higher loss (red) for actions moving it towards the object to reach a play state. Consequently, our curious policy will take actions that navigate the agent closer to the object.

\begin{figure*}[ht]
\begin{center}  
	\includegraphics[width=\linewidth]{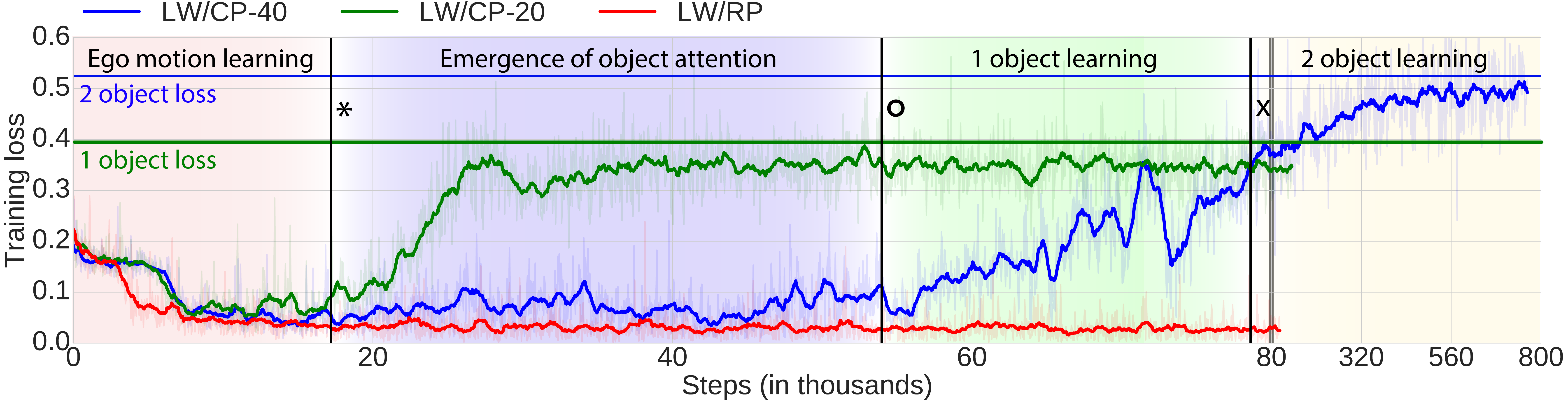} \\
    (a) World model training loss \\
	\begin{tabular}[c]{@{\hspace{-0.1pt}}ccc}
		\includegraphics[width=0.31\linewidth]{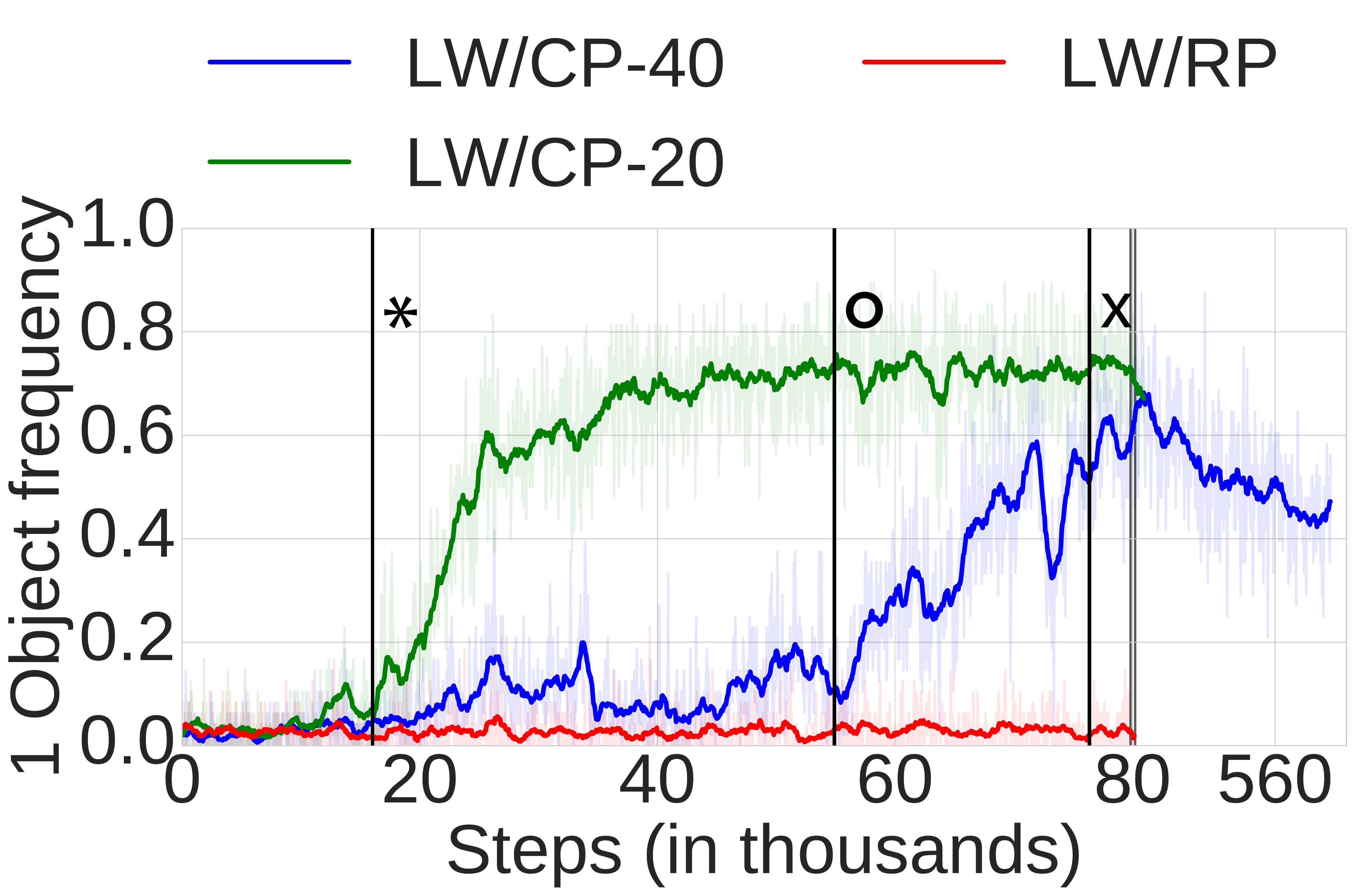}
        &
        \includegraphics[width=0.31\linewidth]{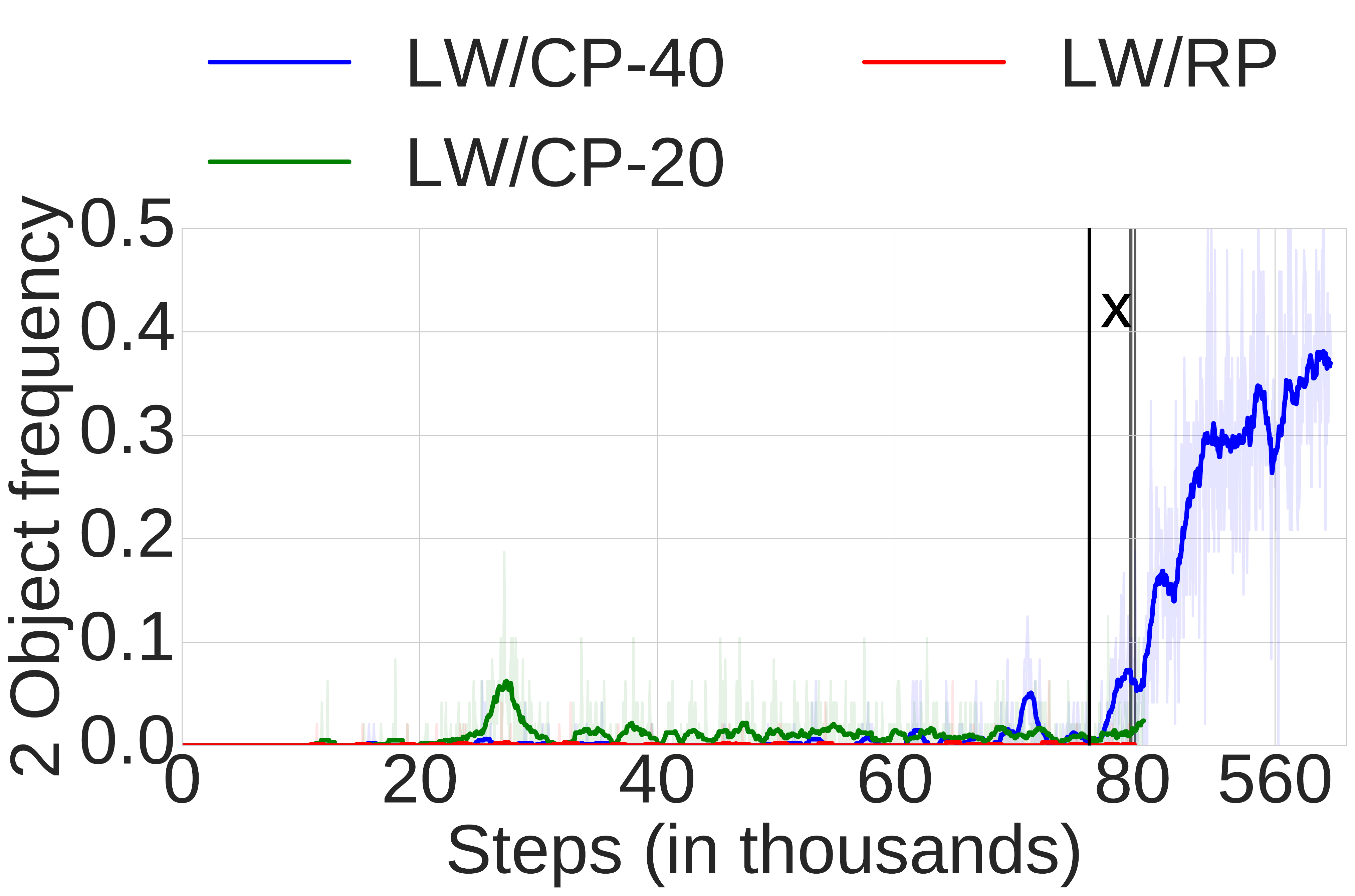} 
        &
        \includegraphics[width=0.31\linewidth]{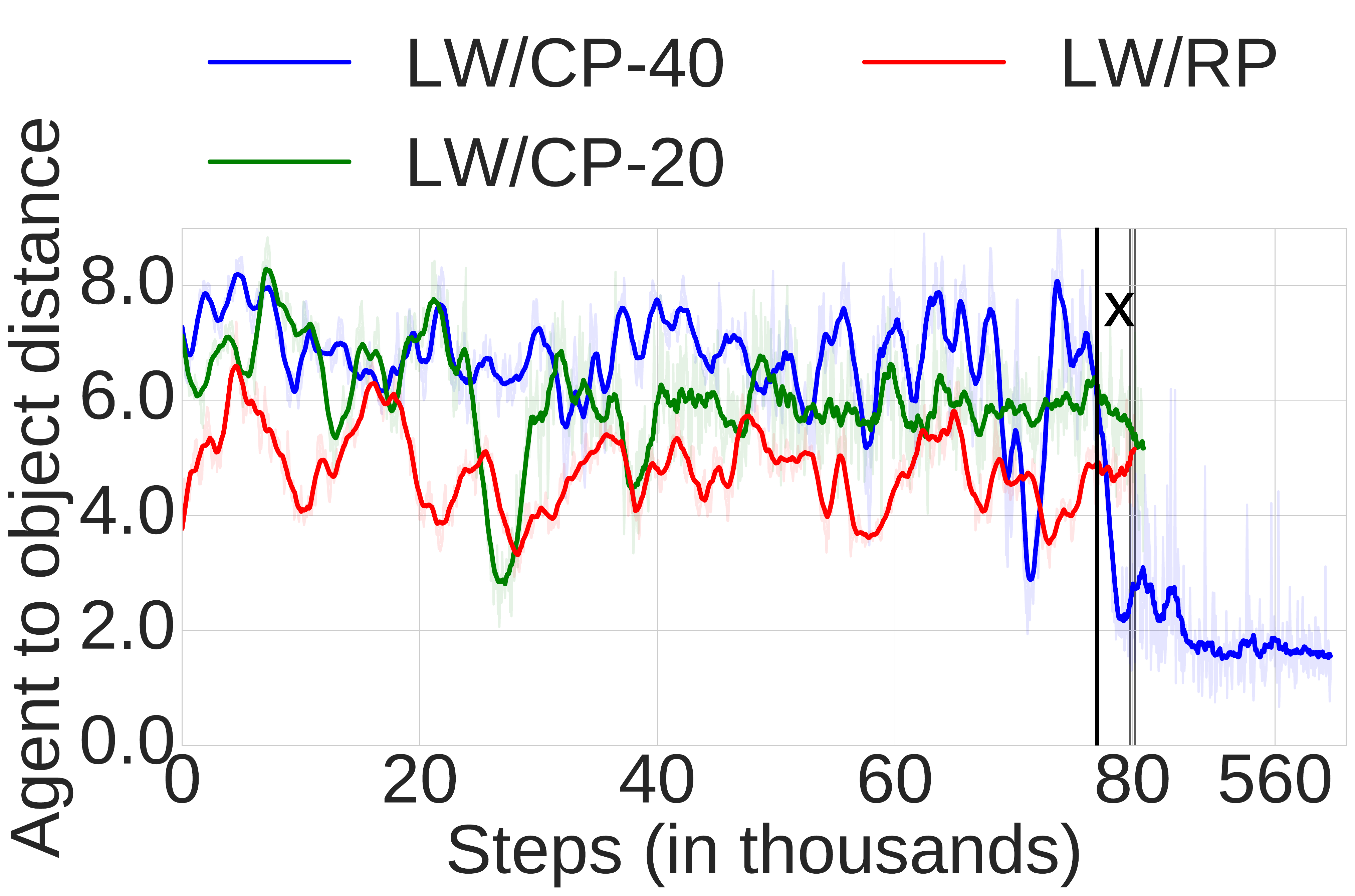}
        \\
        (b) 1 object play frequencies
        &
        (c) 2 object play frequencies
        &
        (d) Object-agent distances 
        \\
	\end{tabular}
\end{center}
\vspace{-0.3cm}
\caption{2 object experiments. Learned world model with curious policy (LW/CP-40) is compared against the same setup but with $T=20$ (LW/CP-20) and a learned world model following a random policy (LW/RP). (a) World model cross-entropy loss during training. (b) Object play state frequency for 1 object in \%. (c) Object play state frequency for 2 objects in \%. (d) Average distance between agent and objects in Unity units.}
\label{fig:2obj_experiments}
\vskip -0.1in
\end{figure*}

\subsection{Emergence of multi-object interactions}

At the beginning of the training of the 2 object experiment we observe similar stages as for the 1 object experiment (Figure \ref{fig:2obj_experiments} (a)). The loss dips as the agent learns to predict its ego motion and rises when its attention shifts towards objects which it then interacts with. This stage is followed by a another loss increase which corresponds to the agent gathering and playing with both objects simultaneously. This is reflected in an increase in 2 object play time (Figure \ref{fig:2obj_experiments} (c)) over 1 object play time (Figure \ref{fig:2obj_experiments} (b)). Consequently, the average distance between the agent and the objects decreases over time as seen in Figure \ref{fig:2obj_experiments} (c). It drops to about 2 units which equals the maximum interaction distance. The LW/RP baseline quickly drops and flattens out. LW/CP-20 with $T=20$ instead of $T=40$ learns to interact with one object but not with two objects simultaneously.

\section{Discussion and Future Work}

We observe that a simple and general intrinsic motivation mechanism based on adversarially antagonizing the loss of a dynamically constructed model of the world allows an agent to stably generate a spectrum of emergent naturalistic behaviors. 
Through self-curricularization in an active learning \cite{settles2011_active} process the agent achieves several ``developmental milestones'' of suitably increasing complexity as it learns to ``play''. 
Starting with random actions, it quickly learns the dynamics of its own ego motion. 
Then, without being given an explicit supervision signal as to the presence or location of an object, it discards ego motion prediction as boring and begins to focus its attention on objects, which are more interesting.
Lastly, when multiple objects are available, it gathers the objects so as to bring them into interaction range of each other.
Throughout, the agent finds its way towards a more challenging data distribution that is at each moment just hard enough to expose the agent to new situations, but still understandable and exploitable by the agent.
This intrinsically motived policy leads to performance gains in its understanding both of the object dynamics, as well as other tasks which the system was not explicitly learning. 

This occurs without any pretrained visual backbone --- the visual system world model was intentionally not initialized with filter weights pretrained on (e.g.) ImageNet classification.  
This result constitutes partial progress in replacing the training of a visual backbone through a task such as large-scale image classification with an interactive self-supervised task and is a proof-of-concept that more complex milestones can be potentially reached while developing an understanding of object categories and physical relations. 

From a machine learning perspective, this combination of spontaneous behavior leading to an improved world model is well suited to designing agents that must act effectively in the many real-world reinforcement learning scenarios in which rewards are sparse or potentially unknown.  
Here, we ultimately seek to develop algorithms that will control autonomous robots that learn to operate in complex unpredictable environments.
From a cognitive science perspective, these results suggest a route toward using intrinsically motivated learning systems to model emergence of spontaneous behavior in young children. 
In this domain, we seek to make computational models that describe key aspects of real infant learning.

However, to truly achieve either of these goals, a variety of limitations of the current work will need to be overcome in future work. 
First, to make the connection to cognitive science more realistic, our environment and agent themselves need to be more realistic. 
On the one hand, better graphics and physics, with more varied and interesting visual objects, will be important to allow better transfer the learned behavior to real-world visuomotor interactions. 
It will also be important to create a properly embodied agent with visible arms and tactile feedback, allowing for more realistic interactions. 
In our current work, which we stress is more a proof-of-principle than a full cognitive model, we fail to address the fact that infants have severely limited mobility and motor control.  
To be able to make realistic predictions for actual infant developmental milestones, it will be especially critical to more realistically model the known developmental trajectory of the motor system. 
In addition, including other animate agents is another way for more complex interactions, but potentially also for better learning through imitation. 

Second, it will be important to improve the reinforcement learning techniques used to better handle more complex interactions beyond those demonstrated here. 
For interactions that are part of a larger experiment, e.g. placing an object on a table or a ramp and then watching it fall, more sophisticated RL policies than those used here are likely to be necessary, with better ability to handle temporally extended reward schedules. 
It will also likely be necessary to use recurrent networks to meet working memory demands in such scenarios.

Third, our \emph{world model} needs to use better representations to improve at predicting such complex interactions. 
Our current approach especially suffers from \emph{degenerate cases} in the inverse dynamical prediction problem --- the problem does not correspond to a well-defined map.   
For example, when given a sequence of images in which an object rests on the ground, the action sequence is under-determined: the agent could have been pushing down on the object, or not.
Though the latent-space approach of \citep{berkeley_mario} is meant in part to ameliorate this issue, we have not yet found an entirely effective solution in this context.
Choices for how to blend components of a generalized inverse dynamical prediction task and the interaction of loss terms seem to be key in resolving this.
Taking these next steps will not only help us to understand how infants learn, but also to develop AI systems that learn without human supervision.

\section*{Acknowledgements}
This work was supported by an Understanding Human Cognition award from the James S. McDonnell foundation (DLKY), a Simons Collaboration on the Global Brain grant (DLKY), a Berry Foundation postdoctoral fellowship (NH), and by hardware donation from the NVIDIA Corporation.

\bibliographystyle{apacite}

\setlength{\bibleftmargin}{.125in}
\setlength{\bibindent}{-\bibleftmargin}

\bibliography{cogsci_2018_learning_to_play}

\end{document}